\documentclass[sigconf,nonacm]{acmart}
\usepackage{algorithm}
\usepackage{algorithmicx}
\usepackage{algpseudocode} 
\usepackage{caption}
\usepackage{subcaption}
\usepackage{csquotes}
\usepackage{amsfonts,amsmath,amsthm}

\usepackage{float} 
\usepackage{placeins} 
\usepackage{multirow} 
\usepackage{tikz}
\usetikzlibrary{positioning}
\usepackage{makecell}
\AtBeginDocument{%
  }

\begin{document}

\title{Efficient GPU Retrieval for Semantic Search}

\author{Dhritiman Das}
\author{Chujie Zheng}
\author{Ronak Kaoshik}
\author{Pratik Dixit}
\author{Vishal Shah}
\affiliation{
 \institution{LinkedIn}
 \city{Mountain View}
 \state{CA}
 \country{USA}}
\email{dhdas@linkedin.com}

\author{Yanbo Li}
\author{Jiahao Xu}
\author{Manika Agarwal}
\author{Chinmay Naik}
\author{Lingyu Zhang}
\affiliation{
 \institution{LinkedIn}
 \city{Mountain View}
 \state{CA}
 \country{USA}}
\email{yanbli@linkedin.com}

\author{Chetan Bhole}
\author{\mbox{Chirag Bhanuprasad Mehta}}
\author{Meng Zheng}
\author{\mbox{Puneet Singh Ahluwalia}}
\author{Shirisha Singh}
\affiliation{
 \institution{LinkedIn}
 \city{Mountain View}
 \state{CA}
 \country{USA}}
\email{cbhole@linkedin.com}

\author{Ping Jin}
\author{Manas Apte}
\author{\mbox{Gokulraj Mohanasundaram}}
\author{Tugrul Bingol}
\authornote{Work done while at LinkedIn.}
\author{\mbox{Raghavan Muthuregunathan}}
\author{Fedor Borisyuk}
\affiliation{
 \institution{LinkedIn}
 \city{Mountain View}
 \state{CA}
 \country{USA}}
\email{pjin@linkedin.com}

\renewcommand{\shortauthors}{Dhritiman Das et al.}


\begin{abstract}
Semantic Search on LinkedIn must retrieve relevant profiles from a corpus of
hundreds of millions in response to natural-language queries such as
\emph{``a fintech founder in Berlin who worked in payments.''} The deployed
relevance policy is bottleneck-oriented: every active non-negotiable facet
must be satisfied, and a pre-existing LLM Graded Relevance (GR) judge
operationalizes this through a fixed min/median aggregation over facet
grades. Cosine similarity instead averages evidence, letting a strong match
on one facet mask failure on another, capping the recall of the first-stage
(L0) retriever.

We present a policy-aligned retrieval framework: embeddings are partitioned
into eight category-supervised segments whose scores follow the same
min/median rule at serving time; for multi-vector retrieval, this segment
score is computed independently per tagged document slot and maximized
across slots. A lightweight single-slot Stage-1 scorer generates high-recall
candidates, while scale-invariant relative-norm gating keeps category
activation consistent across training, evaluation, and serving. On 21K
held-out queries, this representation improves offline relevance over a
matched-capacity baseline, with gains broadly distributed across facet
combinations.

We serve this framework with a two-stage GPU architecture: an FP8 coarse
ranker scores the full corpus, increasing per-shard capacity by 71\% and Stage-1 matmul throughput by 36\%, then an FP16
stage exactly re-ranks an oversampled candidate set, recovering
99.6--99.8\% of full-FP16 recall at over 500 QPS per shard replica. In a
member-randomized A/B test, exploratory-query Precision@10 under the
unchanged GR judge rises from 63.7\% to 79.0\% and navigational
Precision@1 from 65.5\% to 74.7\%, with a blinded human evaluation
independently confirming the Precision@10 gain.
\end{abstract}

\maketitle
\renewcommand{\thefootnote}{\fnsymbol{footnote}}

\section{Introduction} Modern search engines must handle expressive, natural-language queries that describe multi-faceted entities. Rather than matching keywords alone, a system must interpret full-sentence descriptions containing spatial, temporal, and categorical constraints. We study \emph{Semantic People Search} (SPS) on LinkedIn's professional network---for example, \emph{``a fintech founder in Berlin who worked in payments''}---as a case study in high-quality, efficient first-stage GPU retrieval at scale. SPS uses a retrieval funnel: the L0 stage filters hundreds of millions of profiles to a few thousand candidates through embedding-based retrieval (EBR), after which downstream models re-rank the candidates. L0 recall therefore bounds downstream quality. This stage is unusually challenging because the frozen product relevance policy is \emph{conjunctive}: a profile must satisfy all non-negotiable attributes. A pre-existing LLM Graded Relevance (GR) judge operationalizes it through fixed min/median aggregation over category grades~\cite{semanticsearchlinkedin}. Cosine similarity instead averages evidence across an embedding, allowing a profile that matches several facets strongly but fails a required facet to receive a high score. The scorer is therefore mismatched with this frozen policy, whose held-out agreement is our primary quality objective. We address this mismatch with a \textbf{GR-aligned segment embedding}: a vector partitioned into eight category-supervised segments. The aggregation is applied within each tagged document slot, followed by a maximum across slots (\S\ref{sec:model}). These inherited operators predate this work and remain parameter-free; learning is confined to the embeddings. Because the resulting bottleneck score is incompatible with conventional flat-vector candidate generation, we approximate it with a single-slot, flat-dot-product Stage-1 scorer, empirically high-recall, though not by construction. Scale-invariant relative-norm gating preserves the active-category mask across training, evaluation, and serving, while calibration and full-vector auxiliary objectives stabilize segment-score comparability and the joint embedding geometry. We make this representation deployable through a two-stage GPU architecture. An FP8 stage scores the full corpus, and an FP16 stage exactly re-ranks an oversampled candidate set, recovering 99.6--99.8\% of the full-FP16 baseline's recall while increasing Stage-1 index capacity by 71\% and improving Stage-1 matrix-multiplication throughput by 36\% (\S\ref{sec:online-system}). In evaluation at a shard size of 35M documents, the system sustains over 500 QPS per GPU shard replica. Selection and filtering kernels reduce serving overhead but do not change the retrieval objective or live treatment; we present them after deployment evaluation. Our contributions are: \begin{itemize} \item \textbf{A supervision-aligned retrieval representation.} We introduce a GR-aligned segment multi-vector representation that applies the frozen policy aggregation within each tagged document slot, then maximizes across slots (\S\ref{sec:model}). \item \textbf{Parity-consistent candidate generation.} A single-slot, lightweight Stage-1 scorer provides empirically high-recall candidates for the multi-slot final scorer. Relative-norm gating preserves the active-category mask, and calibration and full-vector losses regularize the representation across training and serving (\S\ref{sec:model}). \item \textbf{A deployable GPU system with live validation.} The FP8/FP16 architecture increases Stage-1 index capacity by 71\%, improves Stage-1 matrix-multiplication throughput by 36\%, and recovers 99.6--99.8\% of full-FP16 recall. A member-randomized A/B test measures gains under the unchanged GR judge \cite{sage2026} on live traffic, while supporting kernel and filtering benchmarks are reported separately (\S\ref{sec:online-system}). \end{itemize} \S\ref{sec:related} reviews related work; \S\ref{sec:sps_problem_def} defines the SPS setting; \S\ref{sec:model} presents the retrieval framework; \S\ref{sec:eval} reports the offline evaluation; and \S\ref{sec:online-system} presents the deployable two-stage system, scaling experiments, supporting optimizations, and deployment lessons.

\section{Related Work}
\label{sec:related}
\paragraph{Embedding-Based Retrieval and Learned Similarity}
Bi-encoders replace lexical overlap with retrieval in a shared vector space~\cite{dpr,sbert}, and LLM-based encoders such as Qwen3 Embedding~\cite{qwen3emb} further improve representation quality. Industrial systems, including Facebook Search~\cite{fbebr}, Que2Search~\cite{que2search}, Que2Engage~\cite{que2engage}, and DLRM-style recommenders~\cite{dlrm}, typically use fixed cosine or dot-product scoring. MoL~\cite{mol-kdd23,mol-www25} instead learns a similarity function paired with a hierarchical indexer, while other work observes that contrastive cosine does not capture absolute relevance~\cite{relevance-filtering}. These approaches nevertheless produce a single scalar score. Our framework instead decomposes the embedding into GR-supervised facet segments and applies the aggregation directly~\cite{sage2026}. Only the representations are learned: the scorer introduces neither a scalar approximation nor trainable aggregation weights. Its first stage is validated empirically rather than assumed to preserve recall (\S\ref{sec:model}).

\paragraph{GPU Serving and Quantization}
Our system extends LiNR~\cite{linr}, LinkedIn's GPU-native retrieval architecture, and adopts ideas of GPU-native Bloom filtering similar to SilverTorch~\cite{silvertorch}. Unlike SilverTorch, our filter kernel computes derived features as a side effect, eliminating a separate index layer. We also analyze how Bloom false positives interact with downstream similarity. Approximate nearest-neighbor methods such as product quantization and HNSW reduce index footprint at the cost of recall~\cite{pq,hnsw}, while GPU implementations include FAISS, CAGRA in cuVS, and single-GPU billion-scale search~\cite{faiss,cagra,bang}. We instead perform exhaustive corpus-wide, model-specific scoring in FP8~\cite{fp8formats,fp8power} and recover near-perfect recall with FP16 re-ranking over an oversampled candidate set, making the framework in \S\ref{sec:model} practical at scale.

\paragraph{Multi-Vector Retrieval and Adaptive Dimensioning}
ColBERT and ColBERTv2~\cite{relevance-colbert,colbertv2} use token-level late interaction (MaxSim) to capture fine-grained correspondences. Our framework uses multiple document vectors at the slot level. A shared encoder, conditioned on distinct prefix tags, produces independent representations of the same document; the final model applies the complete GR scorer within each slot and then takes the maximum slot score (\S\ref{sec:model}). We use the backbone's Matryoshka representations~\cite{mrl} to instantiate separate 512- and 256-dimensional per-slot configurations. Each configuration defines eight equal-width category-supervised segments at its native dimensionality (64 and 32 dimensions per segment, respectively); the 256-dimensional model is not obtained by truncating a trained 512-dimensional segment layout.

\paragraph{Multi-Objective Ranking, Distillation, and LLM Judging}
Multi-task architectures~\cite{ple,esmm} and gradient-balancing methods~\cite{multibalance} are widely used to optimize relevance and engagement in downstream rankers. Our L0 training instead combines contrastive objectives with regularizers designed for retrieval geometry. Standard cross-encoder distillation~\cite{hinton-kd,margin-mse} compresses a judge's output into a single scalar. Our segment representation instead distills the LLM judge's explicit multi-category grading policy~\cite{thomas-llm-judge,pinterest-llm-judge,semanticsearchlinkedin,spsretrieval} into the learned embedding geometry while keeping the policy aggregation parameter-free. This alignment objective determines the design of the GPU serving system.

\section{Problem Definition}
\label{sec:sps_problem_def}
SPS retrieves profiles for natural-language queries and poses three challenges: \textbf{C1}, satisfying every required attribute; \textbf{C2}, matching the category-level relevance definition used by the LLM-judge~\cite{sage2026}; and \textbf{C3}, raising the candidate-set recall ceiling imposed at the first stage.

We address these challenges with tagged multi-vector retrieval (\S\ref{sec:model-mv}) and a GR-aligned segment embedding (\S\ref{sec:model-segment}, C1--C2) that partitions each vector into category-specific subspaces. The embedding encoder is learned; serving applies the inherited min/median aggregation within each slot and the maximum across slots, both parameter-free. The model is served through a two-stage retrieve-then-rank GPU pipeline: a model-specific FP8 coarse retrieval stage over the full corpus, followed by an FP16 scorer over an oversampled candidate set (\S\ref{sec:online-system}).

Queries pass through a three-stage funnel---L0 embedding-based retrieval, L2 MLP ranking, and L3 LLM cross-encoding~\cite{semanticsearchlinkedin}---and we focus on L0 since downstream stages can only re-rank the candidates it retrieves (C3). The baseline L0 retriever is a Siamese bi-encoder with a shared
4B-parameter embedding backbone, comparable in scale and capability to
contemporary models such as ~\cite{qwen3emb}. It encodes
queries online as $q \in \mathbb{R}^{d}$ and profiles offline as
$x_i \in \mathbb{R}^{d}$, with $d = 512$.
We reserve $d_s$ for the dimension of one document slot and $K$ for the number of document slots; hence $d_s\times K$ denotes per-slot dimension by slot count:
\begin{equation}
\label{eq:topk}
\mathrm{Ret}_k(q)
=
\operatorname*{top\text{-}k}_{i}\;
\cos(q,x_i)
=
\operatorname*{top\text{-}k}_{i}\;
\frac{\langle q,x_i\rangle}
{\lVert q\rVert_2\,\lVert x_i\rVert_2}.
\end{equation}
This defines the single-vector baseline; the learned segment and multi-vector representations in \S\ref{sec:model} replace this flat similarity with the GR-aligned scoring structure while retaining the same candidate-retrieval interface. We instantiate the control and treatment as separate native-dimensional configurations from the backbone's Matryoshka representation~\cite{mrl}: the control uses $d=512$, whereas the treatment uses $d_s=256$ for each of $K=3$ slots. The treatment is trained and partitioned directly at 256 dimensions; it is not produced by truncating a trained 512-dimensional segment representation.

Labels come from the LLM GR judge \cite{sage2026} (C2), which grades each query--profile pair on a 0--4 scale across eight categories (Table~\ref{tab:segments}): the final grade takes the minimum over six non-negotiable categories, the median over two negotiable categories, then the minimum of those two group scores. A profile is relevant when this grade is at least 3. This LLM GR product policy \cite{sage2026} predates our work and is neither introduced nor tuned here; alternative aggregation rules would define a different product objective rather than a retrieval-model variant. Our segment model learns embeddings for this scoring structure (\S\ref{sec:model-segment}). Training examples are sampled from historical search impressions. Each positive profile is paired with two hard and two easy negatives, all labeled by the eight category-level GR grades. Labels for both this training set and the independently sampled evaluation set are produced by the same pre-existing SAGE People Search LLM-judge\cite{sage2026}; the judge is neither trained nor tuned in this work (\S\ref{sec:eval}).

\section{Modeling Architecture}
\label{sec:model}

All models share the L0 bi-encoder and are trained with the same in-batch InfoNCE objective~\cite{infonce,dpr}. For a batch of queries $\{q_i\}_{i=1}^{B}$ and a pooled document set $\{x_j\}$ with positive index $p(i)$,
\begin{equation}
\label{eq:infonce}
\mathcal{L}_{\mathrm{NCE}} =-\frac{1}{B}\sum_{i=1}^{B} \log\frac{\exp\!\big(s(q_i,x_{p(i)})/\theta\big)} {\sum_{j}\mathbb{1}[j\notin\mathcal{F}_i] \, \exp\!\big(s(q_i,x_j)/\theta\big)},
\end{equation}
using temperature $\theta$ and a false-negative mask $\mathcal{F}_i$. We first describe two-stage segment retrieval with an empirically high-recall Stage-1 scorer in \S\ref{sec:model-gpuret}, then introduce tagged-slot multi-vector representations in \S\ref{sec:model-mv} and the GR-aligned segment representation in \S\ref{sec:model-segment}. Their composition is the model in \S\ref{sec:model-segmv}.

\subsection{High-Recall Two-Stage Segment Retrieval}
\label{sec:model-gpuret}

\paragraph{Serving constraint}
The model represents each document with $K$ tagged embeddings, where slot~$0$ is also trained as the sole corpus-wide embedding used by Stage~1 (\S\ref{sec:model-mv}). Scoring all $K$ embeddings against the full corpus would multiply both GPU-resident storage and corpus-wide similarity computation by $K$, which is prohibitively expensive at our scale even with FP8 storage. We therefore separate full-corpus candidate generation from candidate-set reranking. Stage~1 scans the corpus using only slot~$0$ embedding and returns an over-fetched candidate set of size $M=rK_{\mathrm{out}}$; Stage~2 gathers all $K$ FP16 slots for those candidates and applies the final multi-vector scorer. We explore configurations such as $K{=}3$, $K_{\mathrm{out}}{=}1000$, and $M{=}8000$, which balance memory capacity against quality; all three parameters, including $M$, remain configurable.

\paragraph{Why a flat full-vector score is insufficient}
Each slot is a $d_s$-dim segment embedding partitioned into eight contiguous GR-supervised segments of width $d_s/8$ (\S\ref{sec:model-segment}); we investigate multiple embedding sizes, where the $512$-dim configuration uses $64$ dimensions per segment and the $256$-dim configuration uses $32$. For $q=[q_0,\ldots,q_7]$ and $x=[x_0,\ldots,x_7]$, the full-vector cosine decomposes as
\begin{equation}
\label{eq:decomp}
\cos(q,x)=\sum_{j=0}^{7}\alpha_j(q,x)\,\cos(q_j,x_j), \qquad
\alpha_j=\frac{\lVert q_j\rVert\,\lVert x_j\rVert}{\lVert q\rVert\,\lVert x\rVert}\ge 0.
\end{equation}
Thus, a flat full-vector dot product or cosine behaves like a weighted sum of facet similarities. The final GR-aligned scorer is instead a bottleneck: it first takes the minimum over the six non-negotiable segments, aggregates the two negotiable segments, and then takes the minimum of the two groups (Eq.~\eqref{eq:graligned}). A profile that strongly matches most facets but fails one non-negotiable facet may therefore rank highly under the flat score but poorly under the final scorer. Since Stage~2 can only reorder documents forwarded by Stage~1, this scorer mismatch can cause irrecoverable candidate loss, independent of the slot-0 limitation discussed below.

\paragraph{GR-aligned Stage-1 scorer}
To align candidate generation with the final bottleneck, an idealized Stage~1 would rank slot~$0$ by the minimum cosine over its active non-negotiable segments:
\begin{equation}
\label{eq:stage1-score}
s_{\mathrm{S1}}(q,x^{(0)})=
\min_{j\in\{0,\ldots,5\}^{\!*}} \cos(q_j,x^{(0)}_j),
\end{equation}
where $^{\!*}$ denotes segments that pass the null gate defined in \S\ref{sec:model-segment}. If every non-negotiable segment is inactive, this target falls back to the negotiable-group aggregation so that ranking remains signal-bearing and consistent with the full GR policy. The non-negotiable minimum is the dominant term of the final score and is an upper bound on the full GR-aligned score, making it substantially better aligned with Stage~2 than a flat sum. Stage~1 approximates this target for efficiency (\S\ref{sec:Two-Stage-Architecture}), yielding the candidate set
\begin{equation}
\label{eq:stage1-candidates}
\mathcal{C}_{M}(q)=\operatorname{top\text{-}}M_i\; s_{\mathrm{S1}}\!\left(q,x_i^{(0)}\right).
\end{equation}

\paragraph{Limits of the slot-0 approximation}
Stage~1 ranks candidates using only slot~$0$, while the final score in Eq.~\eqref{eq:stage2-final} takes the maximum across all $K$ slots. Its candidate set is therefore not recall-preserving by construction: a document scoring poorly on slot~$0$ but strongly on slot~$1$ or~$2$ can be excluded from $\mathcal{C}_M(q)$ at the chosen finite depth and is then unrecoverable by Stage~2. We do not close this gap analytically; $\mathcal{L}_{\mathrm{rec}_0}$ (\S\ref{sec:model-mv}) instead trains slot~$0$ to retain globally informative query--document geometry, and we validate the resulting candidate loss empirically rather than assume it away.

Table~\ref{tab:recall-preservation} evaluates this structural mismatch for the $256\times3$ segment multi-vector configuration through the FP8 Stage-1 / FP16 Stage-2 path. At the over-fetch depth of $M{=}8000$, the flat $256$-dim dot product is used in Stage 1 (\S\ref{sec:Two-Stage-Architecture}), recovers $99.8\%$ of the final FP16 top-$1000$, whereas the non-negotiable minimum reaches $100\%$ recall by $M{\approx}1500$. This experiment validates that the higher over-fetch depth compensates for the flat approximation, supporting the selected M on this evaluation pool; the broader corpus-level FP8 capacity and throughput trade-off is reported separately in \S\ref{sec:fp8-quantization}.

\paragraph{Stage-2 reranking and multi-vector composition}
For each candidate $i\in\mathcal{C}_{M}(q)$, Stage~2 gathers all $K$ FP16
segment slots from host memory. It first applies the complete GR-aligned
segment scorer independently to each slot and then selects the best slot:
\begin{equation}
\label{eq:stage2-final}
s_{\mathrm{comb}}(q,d_i)
=
\max_{0\le k<K}
 s_{\mathrm{GR}}\!\left(q,x_i^{(k)}\right),
\end{equation}
where $s_{\mathrm{GR}}$ is defined in Eq.~\eqref{eq:graligned}. This composition completes the GR aggregation within each slot before selecting the highest-scoring slot; evidence from
different slots cannot compensate for different missing categories. The same score is
used as the document logit during training, in offline evaluation, and in
Stage~2 serving. The final result set is
\begin{equation}
\label{eq:stage2-results}
\mathcal{R}_{K_{\mathrm{out}}}(q)
=
\operatorname{top\text{-}}K_{\mathrm{out}}
\left\{s_{\mathrm{comb}}(q,d_i):i\in\mathcal{C}_{M}(q)\right\}.
\end{equation}
Stage~1 observes only slot~$0$, whereas Stage~2 scores all $K$ slots. Slot~$0$
is therefore trained as the sole corpus-wide Stage~1 representation through
$\mathcal{L}_{\mathrm{rec}_0}$ in \S\ref{sec:model-mv}, while the final
combined contrastive objective uses the same $s_{\mathrm{comb}}$ as
Eq.~\eqref{eq:stage2-final}.

The precision-staging mechanics are summarized in
Appendix~\ref{app:precision-staging}; broader FP8/FP16 capacity, throughput,
and recall trade-offs are reported in \S\ref{sec:fp8-quantization}.

\subsection{Multi-Vector Retrieval}
\label{sec:model-mv}

\paragraph{Representation}
Multi-vector retrieval represents each document with $K$ tagged slot embeddings $x^{(0)},\ldots,x^{(K-1)}\in\mathbb{R}^{d_s}$, while keeping the query as a single vector. Slot~$0$ is the sole document embedding scanned corpus-wide in Stage~1; all $K$ slots participate in Stage~2. Thus, the configuration with $K{=}3$ stores three embeddings per document in total, not an additional global vector plus three slots. The serving layout is detailed in \S\ref{sec:multi-vector-retrieval}.

\paragraph{Tagged slot encodings from distinct prefixes}
The central design choice is how the $K$ slots carry different information. Rather than deriving them as deterministic reprojections of one global vector, each slot is generated from its own prefix: a distinct tag token \texttt{[SLOT$k$]} is prepended to the profile text, and the shared fine-tuned backbone re-encodes the tagged input into $x^{(k)}$. The $K$ slots are therefore independent tagged forward passes over the same profile, allowing each conditioning prefix to emphasize a different aspect. Given a per-slot scorer $s_{\mathrm{slot}}$, the document-level late-interaction score is
\begin{equation}
\label{eq:maxsim}
s_{\mathrm{MV}}(q,d)=\max_{0\le k<K}s_{\mathrm{slot}}\!\left(q,x^{(k)}\right).
\end{equation}
For the plain multi-vector model, $s_{\mathrm{slot}}(q,x^{(k)})=\langle q,x^{(k)}\rangle$. The segment multi-vector model instead applies the complete GR-aligned scorer to each slot and then takes the maximum slot score, as defined in \S\ref{sec:model-segmv}.

\paragraph{Loss design for slot specialization}
Tagging alone does not guarantee slot specialization, so we augment the
contrastive objective with diversity, balancing, and slot-0 recall terms:
\begin{equation}
\label{eq:mvloss}
\mathcal{L}=\mathcal{L}_{\mathrm{NCE}}
+\alpha\,\mathcal{L}_{\mathrm{div}}
+\beta\,\mathcal{L}_{\mathrm{bal}}
+\delta\,\mathcal{L}_{\mathrm{rec}_0}.
\end{equation}
The diversity hinge penalizes highly correlated document slots, the balancing
term prevents collapse onto one dominant slot, and the slot-0 recall objective
preserves a globally informative representation for corpus-wide Stage~1
candidate generation. Appendix~\ref{app:mv-loss-details} gives the complete
loss definitions and training hyperparameters.

\subsection{Segment Embedding}
\label{sec:model-segment}

\paragraph{Learning the GR relevance representation}
We instantiate the GR definition from \S\ref{sec:sps_problem_def} by partitioning a $d_s$-dim vector into eight equal-width segments, each supervised by its category-specific grade. Table~\ref{tab:segments} shows the $512$-dim layout ($8\times64$); the $256$-dim model uses $8\times32$ segments. At inference, every segment is independently L2-normalized when computing cosine similarity. Segment activity is determined by a scale-invariant relative-norm gate designed to preserve the learned active-category behavior across evaluation and serving.

\begin{table}[h]
\centering
\small
\captionsetup{font=small}
\caption{Per-slot GR segment layout: $64$ dimensions per segment at $512$-d and $32$ at $256$-d.}
\label{tab:segments}
\setlength{\tabcolsep}{2.5pt}
\begin{tabular}{@{}clc|clc@{}}
\toprule
Seg. & GR Category & Dims & Seg. & GR Category & Dims \\
\midrule
0 & Temporal & $0{:}64$   & 4 & Education & $256{:}320$ \\
1 & Person Name & $64{:}128$  & 5 & Title/Role & $320{:}384$ \\
2 & Company/Org & $128{:}192$ & 6 & Expertise (negotiable) & $384{:}448$ \\
3 & Location & $192{:}256$ & 7 & Industry (negotiable) & $448{:}512$ \\
\bottomrule
\end{tabular}
\end{table}

\paragraph{Segment-contrastive objective}
Each segment is optimized with a per-segment InfoNCE objective over active
categories. We additionally regularize the concatenated representation with a
full-vector contrastive objective and add null and calibration terms:
\begin{equation}
\label{eq:segloss}
\begin{aligned}
\mathcal{L}_{\mathrm{seg}}
={}&\frac{1}{8}\sum_{k=0}^{7}\mathcal{L}^{(k)}_{\mathrm{NCE}}
+w_{\mathrm{full}}\mathcal{L}_{\mathrm{full}} \\
&+w_{\mathrm{null}}\frac{1}{\max(1,|\mathcal{N}|)}
\sum_{k\in\mathcal{N}}\lVert q_k\rVert
+w_{\mathrm{cal}}\operatorname{Var}_k(\bar c_k).
\end{aligned}
\end{equation}
Here $\mathcal{L}_{\mathrm{full}}$ is Eq.~\eqref{eq:infonce} evaluated on
the concatenated full-vector similarity, $\mathcal{N}$ contains inapplicable
query categories; the null term is therefore zero when $\mathcal{N}$ is empty. Moreover, $\bar c_k$ is the mean positive-pair cosine for segment
$k$ over pairs for which that category is relevant. Per-segment InfoNCE learns
the category-specific subspaces, while $\mathcal{L}_{\mathrm{full}}$ keeps
the concatenated embedding globally coherent instead of leaving its
full-vector geometry unconstrained. The segment retriever does not
rank by that flat score: Stage~1 uses the explicit non-negotiable minimum in
Eq.~\eqref{eq:stage1-score}. The null regularizer suppresses the absolute norms of inactive query segments, which in turn lowers their relative norms under Eq.~\eqref{eq:relative-null-gate}. The calibration term separately reduces systematic scale differences among active segments so their cosine scores are comparable under the GR aggregation.

\begin{equation}
\label{eq:relative-null-gate}
a_k(q)=\mathbb{1}\!\left[
\frac{\lVert q_k\rVert_2}{\lVert q\rVert_2}
\ge \tau_{\mathrm{rel}}
\right].
\end{equation}
The relative gate is used by both the standalone segment scorer and the
combined segmented-slot scorer. Unlike an absolute norm threshold, it is
invariant to whole-vector rescaling and removes a normalization-dependent
source of active-mask drift between training, evaluation, and serving.

The equivalent reference pseudocode is provided in
Appendix~\ref{app:gr-scoring-algorithm}.

\paragraph{GR-aligned aggregation}
The Stage~2 scorer combines active segment cosines according to the rule:
\begin{equation}
\label{eq:graligned}
s_{\mathrm{GR}}(q,x)=
\min\!\left(
\min_{k\in\{0,\ldots,5\}^{\!*}}\cos(q_k,x_k),\;
\operatorname{median}_{k\in\{6,7\}^{\!*}}\cos(q_k,x_k)
\right),
\end{equation}
where $^{\!*}$ denotes active segments under the relative-norm gate in Eq.~\eqref{eq:relative-null-gate}. Here the negotiable-group median is the arithmetic mean of the two cosines when both are active and the sole active cosine otherwise. If one group is fully inactive, the scorer uses the other; if both are inactive, it returns zero. Training updates the encoder representations only; the aggregation itself contains no learned weights.

\paragraph{Training and serving}
We preserve the same active-category mask and GR-aligned scorer across
training, offline evaluation, and Stage~2 serving. Reproducibility details,
including gradient caching, hyperparameter optimization, query slicing, and
numerical precision, are provided in Appendix~\ref{app:segment-training-details}.

\subsubsection{GR-Aligned Segment Multi-Vector Embedding}
\label{sec:model-segmv}
The segment and multi-vector representations compose directly. Each member document is
represented by $K$ tagged slots, and every slot is itself a GR-aligned segment
embedding. Slot~$0$ is additionally trained as the sole corpus-wide Stage~1 representation;
the remaining slots add specialized capacity during candidate reranking.

For each slot $k$, the model first computes the complete GR-aligned score
$s_{\mathrm{GR}}(q,x^{(k)})$ using the same active-category mask and
min/median policy as Eq.~\eqref{eq:graligned}. The document score is then
\begin{equation}
\label{eq:segmv-score}
s_{\mathrm{comb}}(q,d)
=
\max_{0\le k<K}s_{\mathrm{GR}}\!\left(q,x^{(k)}\right).
\end{equation}
This is the same GR-then-max composition used by Stage~2 in Eq.~\eqref{eq:stage2-final}; evidence from different slots cannot compensate for different missing categories. We use
Eq.~\eqref{eq:segmv-score} as the document logit in the primary InfoNCE
objective and preserve the identical scorer in offline evaluation and Stage~2
serving, eliminating train--eval--serve scorer drift.

At serving time, Stage~1 approximates
Eq.~\eqref{eq:stage1-score} using a flat dot-product score on the FP8 copy of slot~$0$ over the full corpus, for high efficiency. Stage~2 gathers all $K$ FP16 segment slots for the retained candidates and
applies Eq.~\eqref{eq:segmv-score}. The slot diversity and balancing regularizers keep the $K$ slots distinct and in use, while $\mathcal{L}_{\mathrm{rec}_0}$ preserves the stand-alone retrieval role of
slot~$0$. We evaluate $K{=}3$ slots at $d_s{=}256$ ($8\times32$ segments) and the larger $d_s{=}512$ configuration; \S\ref{sec:eval} reports the offline results.

\section{Evaluation Design and Offline Retriever Quality}
\label{sec:eval}

\paragraph{Evaluation design}
We evaluate category-uniform L0 relevance (this section), deployment shards correctness/capacity/latency (Sections \S\ref{sec:online-system}--\S\ref{sec:derived_features}), and member-randomized post-launch relevance with only the retriever changed (\S\ref{sec:experiments}). Absolute metrics are not directly comparable across these query populations and output stages.
\paragraph{Offline relevance protocol}
For semantic retrieval quality, we use an EBR$+$GR evaluation harness over the
full corpus and $21{,}237$ unique evaluation queries, sampled
uniformly across the query categories defined by product policy. This
coverage-controlled sampling gives each category comparable influence. Each
model performs exhaustive retrieval over the same corpus using its own scoring
function. For every query, the evaluation retains at least the top-$100$
retrieved profiles for recall computation and the top-$10$ profiles for the
remaining ranking metrics.

The evaluation set is constructed independently of the training data. Training
examples are sampled from historical search impressions, whereas evaluation
queries come from a separate production sample using the uniform category
protocol above. The resulting training and evaluation query sets and
query--profile pairs are disjoint. Both datasets are labeled by the same
pre-existing People Search 8B-LLM judge deployed through the SAGE
policy--precedent evaluation service~\cite{sage2026}; we neither train nor tune
this judge in this work. The unchanged judge applies the same fixed,
retriever-independent product policy independently to every retrieved
query--profile pair $(q,d)$ across the two disjoint datasets. On People Search,
SAGE reports a linear-weighted Cohen's $\kappa$ of $0.73$ between this
evaluator and expert human labels, relative to $0.83$ human--human agreement. These figures support
the judge as a scalable surrogate for expert labeling, although they do not
eliminate systematic judge bias. The resulting final grade is
\[
y_q(d)\in\{0,1,2,3,4\}.
\]
A profile is considered relevant when $y_q(d)\ge3$ and a clearly poor match
when $y_q(d)\le1$. Every query--profile pair entering the reported metrics is
assigned a GR grade; therefore, no unjudged result enters the computation of
RS-NDCG@10, Capped R@10, P@1, P@10, or PMR@10. Judgments are produced
independently and are not reused across model runs.

We report P@1 and P@10, retrieved-set linear-gain NDCG@10
(RS-NDCG@10), Capped R@10, and PMR@10. RS-NDCG@10 orders each model's
own top-$10$ against an ideal ordering of that set; it is not a corpus-level
retrieval metric or a shared-pool comparison. Capped R@10 uses relevant
profiles found within each model's top-$100$ as its reference set, with the
denominator capped at $10$; PMR@10 is the fraction of top-$10$ profiles with
GR grade at most one, so lower is better.
Complete metric definitions are provided in
Appendix~\ref{app:metric-definitions}.

All metrics are computed per query and then macro-averaged over the $21{,}237$
evaluation queries. Repeated executions of the evaluation pipeline produce
aggregate metric variation below $10^{-4}$.

Table~\ref{tab:recall-preservation} separately evaluates two-stage candidate
recall for the  $256\times3$ segment multi-vector configuration on the
same $21{,}237$ evaluation queries over the full corpus through the FP8 Stage-1 / FP16 Stage-2 path.

\paragraph{Stage-1 candidate recall}
The final segment score independently normalizes eight category segments and
applies a bottleneck aggregation, whereas a flat full-vector dot product sums
evidence across all dimensions. A profile can therefore rank highly under the
flat score even when one non-negotiable category is poorly matched. For the
segment model, Stage~1 instead computes the cosine similarity of each active
non-negotiable segment in slot~$0$ and ranks by their minimum
(Eq.~\eqref{eq:stage1-score}); if the non-negotiable group is fully inactive,
it falls back to the negotiable median. Stage~2 then applies the complete FP16
segment-aware multi-vector scorer to the retained candidates.

For Stage-1 depth $M$ and final cutoff $K_{\mathrm{out}}$, let
\begin{equation}
C_M(q)
=
\operatorname{TopM}_{d}\,
s_{\mathrm{S1}}(q,d)
\end{equation}
denote the Stage-1 candidate set, and let
\begin{equation}
R_{K_{\mathrm{out}}}^{\mathrm{FP16}}(q)
=
\operatorname{TopK_{\mathrm{out}}}_{d}\,
s_{\mathrm{comb}}^{\mathrm{FP16}}(q,d)
\end{equation}
denote the reference result set produced by the final FP16 segment-aware
multi-vector scorer. We define Stage-1 candidate recall as
\begin{equation}
\label{eq:stage1-recall}
\operatorname{CandidateRecall}_{M@K_{\mathrm{out}}}(q)
=
\frac{
\left|
C_M(q)
\cap
R_{K_{\mathrm{out}}}^{\mathrm{FP16}}(q)
\right|
}{
K_{\mathrm{out}}
}.
\end{equation}

This metric measures the fraction of the final FP16 top-$K_{\mathrm{out}}$
that survives Stage~1 candidate generation. It is distinct from the
query-level Capped R@10 in Eq.~\eqref{eq:recall-at-10}.

Table~\ref{tab:recall-preservation} reports the macro-average
$\operatorname{CandidateRecall}_{M@1000}$ using trained embeddings and the
actual FP8 Stage-1 / FP16 Stage-2 path. At the over-fetch depth of $M{=}8000$, the flat 256-dimensional dot-product recovers $99.8\%$ of the final FP16 top-$1000$, whereas the non-negotiable segment minimum reaches
$100\%$ by $M{=}1500$. These results show that the flat Stage-1 scorer is empirically high-recall on this evaluation pool and provide empirical support for the selected over-fetch depth. They do not constitute a mathematical recall guarantee for the final multi-slot maximum scorer.

\begin{table}[t]
\centering
\small
\captionsetup{font=small}
\caption{Stage-1 candidate recall for $d_s{=}256$, $K{=}3$.
Fraction of the final FP16 top-$1000$ retained in
the Stage-1 top-$M$; higher is better.}
\label{tab:recall-preservation}
\setlength{\tabcolsep}{3pt}
\begin{tabular}{@{}lccc@{}}
\toprule
Stage-1 depth $M$ &
\makecell{Flat 256-dim\\dot product} &
\makecell{Non-neg. $6{\times}32$-d\\dot product} &
\makecell{Non-neg. segment\\minimum} \\
\midrule
1000 & 44.1\% & 42.0\% & 78.3\% \\
1100 & 47.0\% & --     & 86.1\% \\
1200 & 49.8\% & --     & 94.0\% \\
1500 & 57.5\% & --     & \textbf{100\%} \\
2000 & 68.0\% & 65.1\% & 100\% \\
3000 & 82.7\% & 81.7\% & 100\% \\
5000 & 96.6\% & 97.0\% & --    \\
8000 & 99.8\% & 100\%  & --    \\
\bottomrule
\end{tabular}
\end{table}

\paragraph{Offline relevance and ablations}
Table~\ref{tab:segloss-ablation} isolates the null and calibration components
of Eq.~\eqref{eq:segloss}. All rows include both the per-segment InfoNCE
objective and the full-vector auxiliary InfoNCE objective. Both the standalone
segment model and the combined segmented-slot model use the relative null gate
in Eq.~\eqref{eq:relative-null-gate}; this is an implementation-parity choice
that removes a normalization-dependent source of scorer drift rather than an
ablated factor in Table~\ref{tab:segloss-ablation}.

The null regularizer accounts for most of the observed gain by suppressing
inactive-category interference. The calibration term reduces systematic score
differences across segments, making their cosine similarities more comparable
under the shared min/median rule. Calibration has negligible effect
on RS-NDCG@10 and small positive effects on the other three reported metrics.
We therefore treat it as a refinement rather than the primary source of the
quality gain.

\begin{table}[t]
\centering
\small
\captionsetup{font=small}
\caption{Segment-loss ablation. All rows include per-segment and full-vector
InfoNCE. Capped R@10 uses relevant profiles found within the model's top-$100$
as the reference set, with its denominator capped at $10$; lower PMR@10 is
better.}
\label{tab:segloss-ablation}
\setlength{\tabcolsep}{3pt}
\begin{tabular}{@{}lcccc@{}}
\toprule
Objective & RS-NDCG@10 & Capped R@10 & P@1 & PMR@10 \\
\midrule
Contrastive objectives
    & 0.8800 & 0.8777 & 0.7609 & 0.2787 \\
$+$ Null regularization
    & 0.8918 & 0.8855 & 0.7885 & 0.2595 \\
$+$ Score calibration
    & 0.8917 & 0.8856 & 0.7895 & \textbf{0.2583} \\
\bottomrule
\end{tabular}
\end{table}

\paragraph{Capacity--quality trade-offs}
Table~\ref{tab:mv-eval} reports per-slot dimension $d_s$ by document-slot
count $K$. The standalone $512\times1$ segment model isolates GR-aligned
segmentation without extra document vectors: relative to single-vector L0, it
raises RS-NDCG@10 from $0.8704$ to $0.8820$ and P@10 from $0.5880$ to $0.6377$,
while reducing PMR@10 from $0.2987$ to $0.2809$.

At matched $512\times3$ capacity, adding GR-aligned segmentation to
tagged-slot MaxSim improves all five metrics. For the $256\times3$
setting, it raises P@10 from $0.6379$ to $0.6462$ and reduces PMR@10 from
$0.2806$ to $0.2717$. Across five independent training runs, both three-slot
models significantly outperform single-vector L0 under paired tests over
matched per-query metrics ($p < 0.05$).

The $512\times4$ plain multi-vector model is best offline; however, it and the other $512$-dimensional multi-vector configurations exceed the
co-resident control/treatment serving memory budget, so we observe $256\times3$ combined model to have the best trade-off.
The configurations are not a complete factorial grid: slot count is varied
only for plain multi-vector, and dimension is varied at fixed $K{=}3$ for the
combined model. A fixed-candidate check confirms offline/served Stage-2 scorer
consistency but does not test end-to-end Stage-1 parity.
A fixed-capacity ablation of the multi-vector diversity and balancing objectives
for the $256\times3$ model is provided in
Appendix~\ref{app:mv-component-ablation}.

\begin{table}[t]
\centering
\footnotesize
\captionsetup{font=small}
\caption{Capacity--quality trade-offs on the full corpus. RS-N and C-R abbreviate RS-NDCG@10 and Capped R@10; P1/P10 denote P@1/P@10, and lower PMR is better. $^\dagger$ denotes an offline-only configuration that exceeds the co-resident control/treatment serving memory budget; $^*$ denotes the configuration that fits within the co-resident control/treatment memory budget.}
\label{tab:mv-eval}
\setlength{\tabcolsep}{1.6pt}
\begin{tabular}{@{}lrrrrr@{}}
\toprule
Model (per-slot dim. $\times K$) & RS-N & C-R & P1 & P10 & PMR \\
\midrule
Single L0 ($512\!\times\!1$)
 & 0.8704 & 0.8709 & 0.7093 & 0.5880 & 0.2987 \\
GR segment ($512\!\times\!1$)
 & 0.8820 & 0.8795 & 0.7718 & 0.6377 & 0.2809 \\
\midrule
Tagged MV ($512\!\times\!4$)$^\dagger$
 & \textbf{0.8972} & \textbf{0.8871} & \textbf{0.7952} & \textbf{0.6659} & \textbf{0.2565} \\
Tagged MV ($512\!\times\!3$)$^\dagger$
 & 0.8890 & 0.8817 & 0.7820 & 0.6474 & 0.2699 \\
Tagged MV ($256\!\times\!3$)
 & 0.8824 & 0.8794 & 0.7724 & 0.6379 & 0.2806 \\
\midrule
MV+GR ($512\!\times\!3$)$^\dagger$
 & 0.8917 & 0.8856 & 0.7895 & 0.6622 & 0.2583 \\
MV+GR ($256\!\times\!3$)$^*$
 & 0.8835 & 0.8784 & 0.7731 & 0.6462 & 0.2717 \\
\bottomrule
\end{tabular}
\end{table}

\paragraph{Category-combination analysis}
Appendix~\ref{app:category-combinations} reports all 21 supported GR
combinations and a separate navigational group. The standalone segment model
improves 20 of 21 combinations (macro $+0.2574$); its only regression is the
location--name--title group. At matched three-slot capacity, the combined model
improves 19 of 21 combinations over tagged-slot MaxSim (macro $+0.0514$),
regressing only on the company--name and company--location--name groups. The
gains are therefore broad across single-category and conjunctive queries.

\section{Deployable GPU Retrieval}
\label{sec:online-system}

\subsection{Serving Architecture}
\label{sec:system-overview}
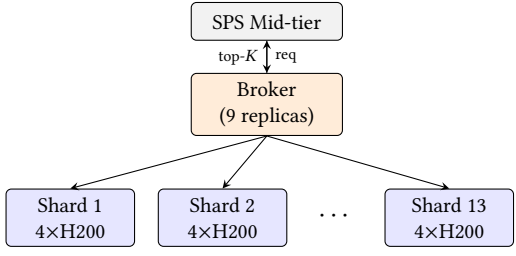
\begin{figure}[t]
\centering
\begin{tikzpicture}[
  font=\small,
  box/.style={draw, rounded corners=2pt, minimum width=2.0cm,
              minimum height=0.5cm, align=center, fill=gray!10},
  gpubox/.style={draw, rounded corners=2pt, minimum width=1.7cm,
                 minimum height=0.5cm, align=center, fill=blue!10},
  arr/.style={->, >=stealth},
]
\node[box] (caller) at (0, 0) {SPS Mid-tier};
\node[box, fill=orange!15] (broker) at (0, -1.1) {Broker\\(9 replicas)};
\draw[arr] (caller) -- (broker) node[midway,right,font=\scriptsize]{req};
\draw[arr] (broker) -- (caller) node[midway,left,font=\scriptsize]{top-$K$};

\node[gpubox] (s1) at (-2.6, -2.6) {Shard 1\\4$\times$H200};
\node[gpubox, right=0.3cm of s1] (s2) {Shard 2\\4$\times$H200};
\node[font=\large, right=0.3cm of s2] (dots) {$\cdots$};
\node[gpubox, right=0.3cm of dots] (s13) {Shard 13\\4$\times$H200};

\draw[arr] (broker.south) -- (s1.north);
\draw[arr] (broker.south) -- (s2.north);
\draw[arr] (broker.south) -- (s13.north);

\end{tikzpicture}
\captionsetup{font=small}
\caption{A sample configuration of SPS GPU-RAR architecture. A broker scatters requests to all shards in parallel, gathers per-shard top-$K$ lists, and merges them.}
\Description{A two-tier retrieval architecture in which an SPS mid-tier calls a replicated broker that scatters requests to 13 H200 GPU shards and merges their top-$K$ results.}
\label{fig:system-overview}
\end{figure}

The GPU-RAR cluster for SPS is organized as a two-tier scatter--gather system. Figure~\ref{fig:system-overview} shows the setup we used for evaluation.

\paragraph{Routing Tier}
A broker layer receives gRPC \texttt{recommend} requests from the SPS mid-tier, \emph{scatters} each request to all inference shards in parallel, \emph{gathers} the per-shard top-$K$ lists and \emph{merges} them into a single ranked result.

\paragraph{Inference Tier}
The corpus is partitioned into disjoint shards using the MD5 hash of the \emph{document key}, with each shard served by one or more single-GPU H200 replicas. Our empirical measurements were conducted on a sample testbed of 13 shards, each with 4 GPU replicas (52 GPU replicas in total per availability zone). Each pod runs an \texttt{inference} container that hosts the index including FP8 embeddings on GPU memory and FP16 embeddings on CPU memory (\S\ref{sec:fp8-quantization}), alongside an \texttt{ingestion} container for index updates from the staging Venice store \cite{venice}.

\paragraph{Index Contents}
Each shard co-hosts a $512$-dimensional single-vector control A/B test variant and the $256\times3$ multi-vector treatment. GPU memory holds their full-corpus Stage-1 FP8 matrices (control and treatment slot~$0$), while host RAM holds the FP16 control vector and all three treatment slots per member. The shard also stores filtering attributes (geographic, company, industry, title, school, language, network flags) and engagement/quality signals retained for downstream ranking outside the fixed GR aggregation (\S\ref{sec:model}).

The 13-shard layout is sized to fit the member corpus at FP8 resolution including multi-vector storage (\S\ref{sec:multi-vector-retrieval}), while the 4-way replication per shard ensures high availability and sufficient throughput headroom for peak traffic.

\subsection{Two-Stage Retrieval}
\label{sec:fp8-quantization}
Large-scale embedding-based retrieval (EBR) engines typically store dense vectors in GPU memory and score queries by matrix multiplication against the full corpus. Quantizing embeddings from FP16 (2 bytes per scalar) to FP8 (1 byte per scalar) halves this footprint. In evaluation, this raised document capacity per shard from 35M to 60M (a 71\% increase) and improved matmul throughput by $\approx\!36\%$ on Hopper-class hardware.

FP8 \texttt{e4m3} has a 3-bit mantissa and a limited dynamic range ($\pm448$). In a single-pass matmul, rounding can reorder near-neighbors, reducing top-1000 recall to 89--93\% across tests with 15M--50M documents. We therefore combine high-speed FP8 coarse filtering over the full corpus with high-fidelity FP16 reranking of an oversampled candidate set.

\subsubsection{Architecture}
\label{sec:Two-Stage-Architecture}
Let query $\mathbf{q}$ score document $i$, whose FP8 Stage-1 slot $\tilde{\mathbf{x}}^{(0)}_i$ resides on GPU while $K_s$ FP16 slots $\{\mathbf{x}^{(k)}_i\}_{k=0}^{K_s-1}$ reside in host RAM; $K_s$ is the slot count and $K$ the requested result depth. FP16-to-FP8 conversion is internal. Algorithm~\ref{alg:two-pass} in Appendix~\ref{app:two-pass-algorithm} gives the complete procedure. The pipeline is implemented as a PyTorch model with custom CUDA kernels integrated for performance-critical operations detailed in  \S\ref{sec:fast-topk}.

\textit{Stage 1: Single slot coarse FP8 retrieval.}
For efficiency, the model scans only slot~0, ranking by flat 256 dim dot-product in FP8. Attribute filters precede top-$M$ selection, with $M=rK$ ($K{=}1000$, $r{=}8$, with $M{=}8000$). This is an empirically high-recall candidate generator, not a recall-preserving bound for the final multi-slot maximum.

\textit{Stage 2: GR-aligned FP16 multi-slot reranking.}
For every retained candidate, Stage~2 gathers all FP16 slots and computes the score
$s_i=\max_k s_{\mathrm{GR}}(\mathbf{q},\mathbf{x}^{(k)}_i)$
(as in Eq.~\eqref{eq:segmv-score}); the final result is the top $K$ candidates by $s_i$. Candidate IDs return to host for slot gathering, and only the $M$-candidate FP16 tensor is transferred back to GPU. Consequently, transfer and exact scoring scale with rerank depth $M$ rather than corpus size $N$. This path is identical to the Stage-1/Stage-2 design in \S\ref{sec:model}; we choose to use a lightweight dot-product scorer in Stage 1 for inference efficiency and rerank a higher top-M ($M{=}8000$) to retain 99.8\% recall.

\subsubsection{Empirical Results}
Table~\ref{tab:two-pass-results} summarizes the trade-off. Single-pass FP8 loses 5--11\% recall relative to exact FP16, whereas an all-FP16 GPU index cannot support the larger shard. The two-pass design recovers 99.6--99.8\% of exact-FP16 recall while supporting 60M documents. The $1.36\times$ value is Stage-1 matmul throughput, not end-to-end throughput: it excludes the host gather, multi-slot scoring, and final top-$K$ selection.  QPS and latency are reported in \S\ref{sec:multi-vector-retrieval}. FP16 storage moves to host RAM, increasing allocation from $\approx\!30$\,GB to 150\,GB per shard; GPU retains the Stage-1 FP8 tensors, with only treatment slot~0 used in Stage~1. This host-RAM budget for multi-vector storage caps the practical shard size at ~35M documents, even though FP8 alone could support 60M for a single vector.

\begin{table}[t]
\centering
\footnotesize
\captionsetup{font=small}
\caption{Single-pass vs. two-pass FP8 retrieval on a sample evaluation configuration ($K{=}1000$; recall relative to exact FP16).}
\label{tab:two-pass-results}
\setlength{\tabcolsep}{2.4pt}
\begin{tabular}{@{}lccc@{}}
\toprule
Metric & FP16 & FP8 & \makecell{Two-pass\\FP8+FP16} \\
\midrule
GPU bytes/scalar    & 2 B & 1 B & 1 B (GPU) \\
Recall@1000 vs. FP16 & 100\% & 89--93\% & 99.6--99.8\% \\
Stage-1 matmul      & $1.0\times$ & $1.36\times$ & $\approx1.36\times$ \\
Host RAM/shard      & 30 GB & 30 GB & 150 GB \\
Documents/shard     & 35M & 60M & 60M \\
\bottomrule
\end{tabular}
\end{table}

\subsection{Multi-Vector Reranking}
\label{sec:multi-vector-retrieval}
Single-vector retrieval loses aspect-level correspondence during the inner-product match; multi-vector late interaction (e.g., ColBERT~\cite{relevance-colbert,colbertv2}) restores it with multiple document slots. We apply this refinement only to the small FP16 re-ranked candidate set, requiring no request-schema changes.

\paragraph{Scoring and serving}
The query remains one vector $q$, while document $d_i$ has $K$ tagged segment slots $x_i^{(0)},\dots,x_i^{(K-1)}\in\mathbb{R}^{d_s}$ (\S\ref{sec:model-mv}). Stage~1 scans the FP8 copy of slot~$0$ with FP8 dot-product; Stage~2 gathers all $K$ FP16 slots for the retained candidates, evaluates the complete GR scorer within each slot, and takes the maximum (Eq.~\eqref{eq:segmv-score}). The CPU supplies only the $S \times d_s$ block per slot, and the GPU computes batched segment dot products followed by the parameter-free min/median and maximum reductions, focusing on $K{=}3$, $d_s{=}256$ for its better memory--quality trade-off. Transfer therefore scales with re-rank depth $S$, rather than corpus size $N$; \texttt{first\_stage\_topk} controls the latency--recall trade-off.

\paragraph{Performance}
At 400 QPS, increasing $S$ from $2{,}000$ to $20{,}000$ raises P95 latency from 72.8 to 158.7\,ms. Under a P95 target of 150\,ms and a gRPC error-rate limit of 0.5\%, sustainable capacity ranges from 600 to 300 QPS. The complete sweep is reported in Appendix~\ref{app:mvr-full-sweep}.

\subsection{Live-Traffic Relevance Evaluation}
\label{sec:experiments}
The live A/B test evaluates final results under natural traffic, unlike Section~\ref{sec:eval}'s category-uniform L0 study. From April through June 2026, eligible English-speaking members were assigned at the member level to control or treatment buckets, with treatment ramped to 50\%. Query decoding and L2/L3 ranking remained unchanged, so the only experimental change was the L0 retriever.

Navigational queries seek one or a small number of specific profiles (for example, the person holding a named organizational role), whereas exploratory queries admit many relevant profiles. To reduce query-mix bias, the comparison uses high-frequency query strings observed in both buckets rather than disjoint arm-specific query sets. The online precision metrics are not click or engagement rates: they are computed by applying the unchanged SAGE relevance judge \cite{sage2026} to query--profile results sampled from live traffic. This judge was neither trained nor tuned for the experiment; its prior validation and our supplementary human robustness check are described in \S\ref{sec:eval} and Appendix~\ref{app:online-human-eval}.

The judged results show exploratory P@10 increasing from 63.7\% to 79.0\%, an absolute lift of 15.3 percentage points and a relative lift of 24.0\%. Navigational P@1 increases from 65.5\% to 74.7\%, an absolute lift of 9.2 points and a relative lift of 14.0\%. The A/B experimentation platform reports $p<0.03$ for both differences. A supplementary blinded human evaluation ($n=50$, Appendix~\ref{app:online-human-eval}) independently confirms a significant Human P@10 gain (+6.0 points, $p=0.004$); it also finds a directionally consistent Human P@1 gain (+8.0 points) that does not reach significance at this sample size ($p=0.424$). These online lifts exceed the matched-capacity offline gains in \S\ref{sec:eval} (P@10 0.6379$\to$0.6462), because live traffic is not category-uniform and skews toward query types where the treatment gains most. Member-engagement guardrails, including search-session measures, were monitored separately; no regression was observed.

\subsection{Supporting GPU Optimizations}
\label{sec:supporting-gpu-optimizations}
These optimizations reduce serving overhead but do not change the retrieval objective or the treatment evaluated in the live A/B test.

\paragraph{Exact top-$K$.}
\label{sec:fast-topk}
Blitz replaces \texttt{torch.topk} with an exact, sorted, two-pass FP16 radix-select CUDA kernel. Across the four target H200 configurations ($K{=}1000$, batches 1 and 32, and 50M--100M scores), it is $2.9$--$5.6\times$ faster than \texttt{torch.topk} while preserving 100\% value overlap; index differences occur only among tied scores. Appendix~\ref{app:fast-topk} gives the implementation, full scale sweep, and fallback behavior.

\paragraph{Network filtering.}
\label{sec:bloom-filter-optimization}
Network-distance filtering uses a query-side Bloom filter for large second-degree connection sets and exact matching for first-degree connections. This keeps the GPU index compact and stateless; implementation details and false-positive trade-offs are reported in Appendix~\ref{sec:fpr_bloomfilter}.

\paragraph{Derived features.}
\label{sec:derived_features}
The dual-role GPU filter computes both network-derived indicators in one retrieval call, replacing three parallel calls while reusing candidate fetches. Appendix~\ref{sec:appendix_derived_features} reports the complete quality and load-test results.

\subsection{Lessons Learned}
\label{sec:deployment-lessons}
This work yields four practical lessons. First, in evaluation, the flat scorer needed $M=8000$ for 99.8\% recall, whereas the costlier segment-aware minimum reached 100\% by $M=1500$; the lesson is to keep the cheap flat scorer and compensate with larger depth. Second, FP8 suits coarse scanning, not final ranking: observed single-pass recall was 89--93\%, while candidate-only FP16 reranking recovered 99.6--99.8\%. Third, offline quality is bounded by the memory budget: a $512\times4$ model led offline but proved infeasible to serve, whereas $256\times3$ balanced quality and memory. Finally, FastTopK, Bloom filtering, and derived features reduce serving overhead without changing the relevance scorer or A/B treatment.

\section{Conclusion}
We presented an end-to-end retrieval design that aligns the learned representation and its GPU serving path with the product GR definition. Category-supervised segments preserve required facets, tagged slots capture complementary profile aspects, and a dedicated slot-0 scorer supports full-corpus candidate generation. On 21,237 disjoint offline queries, category-uniform, matched-capacity, and per-category evaluations show that the gains are not explained solely by additional vector capacity; a blinded 50-query evaluation independently confirms the P@10 improvement, though the corresponding P@1 gain is directional rather than significant at this sample size. The live A/B test (\S\ref{sec:experiments}) confirms these offline gains translate into online metric lifts.

The FP8/FP16 path increases per-shard capacity by 71\%, recovers 99.6--99.8\% of full-FP16 recall, and sustains over 500 QPS per GPU shard replica, demonstrating at-scale practicality.

\bibliographystyle{ACM-Reference-Format}
\bibliography{sample-base}

\clearpage\appendix
\setlength{\textfloatsep}{8pt plus 2pt minus 2pt}
\setlength{\floatsep}{6pt plus 2pt minus 2pt}
\setlength{\intextsep}{6pt plus 2pt minus 2pt}

\section{Modeling and Serving Details}

\subsection{Precision-staging implementation}
\label{app:precision-staging}
Stage~1 keeps only the FP8 copy of slot~$0$ resident on GPU and computes a single flat FP8 dot product $\langle \mathbf{q}, \tilde{\mathbf{x}}^{(0)} \rangle$ against the full corpus. The coarse FP8 matrix multiplication also carries the padded and filtered $-\infty$ mask. Stage~1 ranks candidates directly by this flat inner product, as a lightweight serving-time approximation of the segment-aligned scorer defined in Eq.~\eqref{eq:stage1-score}; no per-segment cosine decomposition or min/median aggregation is applied at this stage. For the $M$ retained candidates, Stage~2 gathers all $K$ FP16 slots from host memory and transfers the candidate tensor to GPU for the complete segment-aware multi-vector aggregation in Eq.~\eqref{eq:segmv-score}. This flat slot-0 candidate generator is empirically high-recall but is not recall-preserving by construction for the final maximum over all slots.

\subsection{Reference two-pass ranking algorithm}
\label{app:two-pass-algorithm}
\begin{algorithm}[H]
\small
\captionsetup{font=small}
\caption{Two-Pass FP8/FP16 Ranking}
\label{alg:two-pass}
\begin{algorithmic}[1]
\Require query $\mathbf{q}$, FP8 slot-0 corpus $\{\tilde{\mathbf{x}}^{(0)}_i\}$ on GPU, all FP16 slots $\{\mathbf{x}^{(k)}_i\}$ in RAM, filter $P$, result depth $K$, ratio $r$
\State $\tilde{s}_i \gets \langle \mathbf{q}, \tilde{\mathbf{x}}^{(0)}_i\rangle$ for all $i$ \Comment{flat FP8 dot product (Stage 1)}
\State $\mathcal{V} \gets \{\, i : P(i) \text{ holds} \,\}$ \Comment{attribute-based filtering}
\State $\mathcal{C} \gets \operatorname*{top\text{-}}(rK)$ of $\tilde{s}$ over $\mathcal{V}$ \Comment{oversampled candidates}
\State gather all FP16 slots $\{\mathbf{x}^{(k)}_i : i \in \mathcal{C}\}$ from RAM
\State $s_i \gets \max_k s_{\mathrm{GR}}(\mathbf{q},\mathbf{x}^{(k)}_i)$ for $i \in \mathcal{C}$
\State \Return $\operatorname*{top\text{-}}K$ of $\{s_i : i \in \mathcal{C}\}$
\end{algorithmic}
\end{algorithm}

\subsection{Multi-vector regularization details}
\label{app:mv-loss-details}
For a batch of $B$ positive query--document pairs, writing $q_i$ for a query
and $x_i^{(0)},\ldots,x_i^{(K-1)}$ for its document slots, the auxiliary terms
in Eq.~\eqref{eq:mvloss} are
\begin{align}
\mathcal{L}_{\mathrm{div}}
&=\frac{1}{B K(K-1)}\sum_{i=1}^{B}\sum_{\substack{0\le j,k<K\\j\ne k}}
\left[\max\!\left(0,\left|\cos(x_i^{(j)},x_i^{(k)})\right|-m\right)\right]^2, \\
\mathcal{L}_{\mathrm{bal}}
&=\sum_{k=0}^{K-1}\left(\bar p_k-\frac{1}{K}\right)^2, \\
\mathcal{L}_{\mathrm{rec}_0}
&=-\frac{1}{B}\sum_{i=1}^{B}
\log\frac{\exp(\langle q_i,x_i^{(0)}\rangle/\theta_0)}
{\sum_j\exp(\langle q_i,x_j^{(0)}\rangle/\theta_0)},
\end{align}
where
\begin{equation}
\bar p_k=\frac{1}{B}\sum_i
\frac{\exp(\langle q_i,x_i^{(k)}\rangle/\tau_{\mathrm{bal}})}
{\sum_{k'}\exp(\langle q_i,x_i^{(k')}\rangle/\tau_{\mathrm{bal}})}.
\end{equation}
The diversity term is a squared hinge on absolute pairwise slot cosine above
margin $m$; the balancing term drives batch-mean slot usage toward $1/K$; and
$\mathcal{L}_{\mathrm{rec}_0}$ is an in-batch InfoNCE objective applied to
slot~$0$. For the configuration evaluated in our experiments, the multi-vector regularizer coefficients are set to $\alpha{=}0.01$, $\beta{=}0.01$, and $\delta{=}0.01$ for diversity, balancing, and slot-0 recall, respectively. The remaining hyperparameters are $m{=}0.3$, $\tau_{\mathrm{bal}}{=}0.05$, and $\theta_0{=}0.05$.

\subsection{Training-objective component analysis}
\label{app:mv-component-ablation}
To isolate the effects of the two slot-specialization regularizers without
changing the candidate-generation path, every row in
Table~\ref{tab:mv-component-ablation} uses the  
$256\times3$ GR-aligned segment multi-vector architecture, tagged slots,
training data, optimization budget, and evaluation protocol. Every row also
uses the same shared base objectives: the primary multi-vector contrastive
objective, the per-segment and full-vector auxiliary InfoNCE objectives, null
regularization, score calibration, and the slot-0 recall objective
$\mathcal{L}_{\mathrm{rec}_0}$. Consequently, every configuration trains a
standalone slot~0 for Stage-1 candidate generation. The fixed GR aggregation is identical in every row and is not an ablation variable; the comparison varies only
the diversity and balancing regularizers.

\begin{table*}[t]
\centering
\small
\captionsetup{font=small}
\caption{Training-objective component analysis for the
$256\times3$ configuration. Every row includes all shared base objectives
listed above, including the contrastive objectives and slot-0 recall loss.
Only $\mathcal{L}_{\mathrm{div}}$ and $\mathcal{L}_{\mathrm{bal}}$ are
ablated; \textbf{Y} indicates an enabled regularizer. Lower PMR@10 is better.}
\label{tab:mv-component-ablation}
\setlength{\tabcolsep}{3pt}
\begin{tabular}{@{}lccccccc@{}}
\toprule
Configuration &
$\mathcal{L}_{\mathrm{div}}$ &
$\mathcal{L}_{\mathrm{bal}}$ &
RS-NDCG@10 & Capped R@10 & P@1 & P@10 & PMR@10 \\
\midrule
Base objectives
 & -- & -- & \textit{0.8804} & \textit{0.8775} & \textit{0.7558} & \textit{0.6244} & \textit{0.278} \\
Base $+$ diversity
 & \textbf{Y} & -- & \textit{0.8832} & \textit{0.8777} & \textit{0.7730} & \textit{0.6462} & \textit{0.2776} \\
Base $+$ balancing
 & -- & \textbf{Y} & \textit{0.8820} & \textit{0.8785} & \textit{0.7718} & \textit{0.6377} & \textit{0.2809} \\
Full model
 & \textbf{Y} & \textbf{Y} &
 0.8835 & 0.8784 & 0.7731 & 0.6462 & 0.2717 \\
\bottomrule
\end{tabular}
\end{table*}

Relative to the shared-base configuration, diversity regularization provides
most of the top-rank precision gain, increasing P@1 from $0.7558$ to $0.7730$
and P@10 from $0.6244$ to $0.6462$. Balancing alone gives the highest
Capped R@10 ($0.8785$), but slightly worsens PMR@10 ($0.2809$ versus
$0.278$). Combining the two regularizers preserves the diversity-only P@10
of $0.6462$ and reduces PMR@10 to $0.2717$, the best value in the ablation,
while also achieving the highest RS-NDCG@10 ($0.8835$) and P@1 ($0.7731$).
These results suggest complementary roles: diversity drives most of the
precision improvement, whereas balancing is most useful alongside diversity
for reducing poor matches.


\subsection{Reference GR-aligned scorer}
\label{app:gr-scoring-algorithm}
\begin{algorithm}[H]
\small
\captionsetup{font=small}
\caption{GR-aligned segment scoring with relative null gating}
\label{alg:graligned}
\begin{algorithmic}[1]
\Require query $q$, document $x$, fixed relative threshold $\tau_{\mathrm{rel}}$
\State partition $q\!\to\![q_0,\ldots,q_7]$, $x\!\to\![x_0,\ldots,x_7]$
       ($q_k,x_k\in\mathbb{R}^{d_s/8}$)
\For{$k=0$ \textbf{to} $7$}
  \State $a_k \gets \mathbb{1}[\lVert q_k\rVert_2/\lVert q\rVert_2 \ge \tau_{\mathrm{rel}}]$
  \State $c_k \gets \cos(q_k,x_k)$
\EndFor
\State $s_{\mathrm{nn}} \gets \min\{c_k : a_k,\ k\in\{0,\ldots,5\}\}$
\State $s_{\mathrm{negotiable}} \gets \operatorname{median}\{c_k : a_k,\ k\in\{6,7\}\}$ \Comment{mean of two, or the sole active value}
\State \Return $\min(s_{\mathrm{nn}},s_{\mathrm{negotiable}})$, skipping any fully inactive group
\end{algorithmic}
\end{algorithm}

\subsection{Segment training and numerical settings}
\label{app:segment-training-details}
We train on $1{,}034{,}554$ GR-labeled $(\text{query},\text{positive},\text{negative})$ triples constructed from historical search impressions. Both query and profile text are truncated to
$1{,}500$ tokens. The query input consists of \enquote{Instruct: Given a web search query,
retrieve relevant passages that answer the query}, followed by
\enquote{Query:} on a new line. We fine-tune a 4B-parameter embedding backbone, comparable in scale and
capability to contemporary open-source models such as ~\cite{qwen3emb}, for one epoch using BF16 precision. For the experimental configuration reported here, training uses an effective batch size of $256$, an encoding mini-batch size of $32$, a learning rate of $9.931 \times 10^{-6}$, and a warmup ratio of $0.03677$, together with gradient checkpointing and gradient caching~\cite{gradcache}. The primary and full-vector contrastive temperatures are set to $0.02008$ and $0.05$, respectively. The loss weights are $w_{\mathrm{null}}{=}0.2689$, $w_{\mathrm{cal}}{=}2.074$, and $w_{\mathrm{full}}{=}0.5$. Both the positive-score cutoff and the false-negative masking threshold are set to $2.5$, corresponding to an integer GR grade of at least $3$. These values were selected through Optuna-based hyperparameter optimization~\cite{optuna}; the final evaluation dataset was held out from the optimization process.

Evaluation and serving use the query at the native dimensionality of the selected configuration.
The 512- and 256-dimensional configurations define their segment boundaries independently rather than truncating one segmented representation into the other.
No additional whole-vector normalization is applied, preserving the segment components used by the relative null gate. Segment
norms are computed in float32 for numerical stability under FP16 reranking.

\FloatBarrier

\section{Offline Evaluation Metric Definitions}
\label{app:metric-definitions}
For completeness, we give the per-query definitions used by the offline
evaluation before macro-averaging over the evaluation query set.

Let $T_k^m(q)$ denote the first $k$ profiles returned by model $m$ for query
$q$, and define the relevant subset at depth $k$ as
\begin{equation}
\label{eq:relevant-set}
\operatorname{Rel}_k^m(q)
=
\left\{
d\in T_k^m(q):y_q(d)\ge3
\right\}.
\end{equation}

\paragraph{Precision}
Precision measures the fraction of returned profiles judged relevant. For
cutoff $k\in\{1,10\}$, we define
\begin{equation}
\label{eq:precision}
\operatorname{P@}k_m(q)
=
\frac{
\left|\operatorname{Rel}_k^m(q)\right|
}{
\left|T_k^m(q)\right|
}.
\end{equation}
When a query returns at least $k$ profiles, as is normally the case in this
evaluation, $|T_k^m(q)|=k$. In particular,
\begin{equation}
\operatorname{P@10}_m(q)
=
\frac{
\left|\operatorname{Rel}_{10}^m(q)\right|
}{
10
}.
\end{equation}

\paragraph{Capped Recall@10}
Recall@10 is computed relative to the relevant profiles found within the
model's first $100$ retrieved results. Let
\begin{equation}
\label{eq:top100-relevant}
R_{100}^m(q)
=
\left|\operatorname{Rel}_{100}^m(q)\right|.
\end{equation}
We define
\begin{equation}
\label{eq:recall-at-10}
\operatorname{CappedR@10}_m(q)
=
\begin{cases}
\displaystyle
\frac{
\left|\operatorname{Rel}_{10}^m(q)\right|
}{
\min\!\left(R_{100}^m(q),10\right)
},
&
R_{100}^m(q)>0,
\\[10pt]
0,
&
R_{100}^m(q)=0.
\end{cases}
\end{equation}
This is a depth-$100$, top-$10$-capped recall metric rather than classical
corpus-level recall. It measures how many of the up to ten relevant profiles
found within the model's top-$100$ are concentrated in its top-$10$.

Capped R@10 and P@10 use the same relevance threshold and numerator, but
different denominators. P@10 divides by the number of returned top-$10$
positions, whereas Capped R@10 divides by the number of relevant profiles
found within the top-$100$, capped at ten. The two metrics therefore need not
be numerically equal.

\paragraph{Retrieved-set graded NDCG}
Let $d_i^m$ denote the profile returned by model $m$ at rank $i$. Following
the evaluation implementation, DCG uses the GR score directly as a linear
gain:
\begin{equation}
\label{eq:dcg}
\operatorname{DCG@10}_m(q)
=
\sum_{i=1}^{|T_{10}^m(q)|}
\frac{
y_q(d_i^m)
}{
\log_2(i+1)
}.
\end{equation}
The ideal DCG is obtained by sorting the GR scores of the same retrieved
top-$10$ profiles in descending order. Let
\[
y_{q,1}^{m,\downarrow}
\ge
y_{q,2}^{m,\downarrow}
\ge
\cdots
\ge
y_{q,|T_{10}^m(q)|}^{m,\downarrow}
\]
denote these sorted scores. Then
\begin{equation}
\label{eq:idcg}
\operatorname{IDCG@10}_m(q)
=
\sum_{i=1}^{|T_{10}^m(q)|}
\frac{
y_{q,i}^{m,\downarrow}
}{
\log_2(i+1)
}.
\end{equation}
We define
\begin{equation}
\label{eq:ndcg}
\operatorname{RS\text{-}NDCG@10}_m(q)
=
\begin{cases}
\displaystyle
\frac{
\operatorname{DCG@10}_m(q)
}{
\operatorname{IDCG@10}_m(q)
},
&
\operatorname{IDCG@10}_m(q)>0,
\\[10pt]
0,
&
\operatorname{IDCG@10}_m(q)=0.
\end{cases}
\end{equation}
Thus, RS-NDCG@10 measures only the ordering quality of the GR grades within
each model's retrieved top-$10$. Because the ideal list is constructed from
each model's own retrieved set, it does not measure corpus-level retrieval
quality or compare result-set quality against a shared judged pool.

\paragraph{Poor-match rate}
We define the poor-match rate as the fraction of returned top-$10$ profiles
with final GR grade at most one:
\begin{equation}
\label{eq:pmr}
\operatorname{PMR@10}_m(q)
=
\frac{
\displaystyle
\sum_{d\in T_{10}^m(q)}
\mathbb{I}\!\left[y_q(d)\le1\right]
}{
\left|T_{10}^m(q)\right|
},
\end{equation}
where lower is better.

\FloatBarrier

\section{Full Category-Combination Results}
\label{app:category-combinations}

\begin{table*}[t]
\centering
\footnotesize
\captionsetup{font=small}
\caption{Average final GR score (0--4) by query group under the same
category-uniform retriever-quality protocol. $\Delta_{1}$ is Segment only
minus Single-vector L0; $\Delta_{3}$ is MV $+$ segment minus plain MV.}
\label{tab:all-category-combinations}
\setlength{\tabcolsep}{2.5pt}
\begin{tabular}{@{}lrrrrrrr@{}}
\toprule
Query group & $k_g$ & Single & Segment & MV & MV+Seg & $\Delta_1$ & $\Delta_3$ \\
\midrule
education & 813 & 2.8220 & 3.1570 & 3.3137 & 3.3530 & +0.3350 & +0.0393 \\
education, location & 737 & 2.6709 & 2.9511 & 3.1377 & 3.2176 & +0.2802 & +0.0799 \\
education, expertise & 728 & 2.7708 & 3.0494 & 3.1368 & 3.1941 & +0.2786 & +0.0573 \\
education, name & 746 & 1.7058 & 1.8401 & 1.8510 & 1.9211 & +0.1343 & +0.0701 \\
location, name, title & 698 & 2.0010 & 1.9452 & 1.8964 & 1.9487 & -0.0558 & +0.0523 \\
expertise, location & 758 & 2.8825 & 3.0791 & 3.1122 & 3.1601 & +0.1966 & +0.0479 \\
industry & 721 & 3.6110 & 3.7749 & 3.7942 & 3.8148 & +0.1639 & +0.0206 \\
expertise & 749 & 3.2507 & 3.5292 & 3.5466 & 3.6084 & +0.2785 & +0.0618 \\
location, title & 795 & 2.9741 & 3.2456 & 3.2442 & 3.3170 & +0.2715 & +0.0728 \\
title/role & 781 & 3.2628 & 3.6623 & 3.6890 & 3.7342 & +0.3995 & +0.0452 \\
name, title & 744 & 1.9135 & 2.0327 & 2.0415 & 2.1077 & +0.1192 & +0.0662 \\
company, location & 913 & 2.5566 & 2.8396 & 2.8853 & 2.9458 & +0.2830 & +0.0605 \\
expertise, name & 749 & 2.2019 & 2.2561 & 2.3274 & 2.3573 & +0.0542 & +0.0299 \\
company, name & 934 & 1.6276 & 1.7622 & 1.7697 & 1.7258 & +0.1346 & -0.0439 \\
company, expertise & 780 & 2.2925 & 2.7496 & 2.8156 & 2.8692 & +0.4571 & +0.0536 \\
company, title & 897 & 2.3363 & 2.7982 & 2.8787 & 2.9308 & +0.4619 & +0.0521 \\
person name & 3583 & 2.3276 & 2.8883 & 2.8149 & 2.9108 & +0.5607 & +0.0959 \\
location & 738 & 3.3613 & 3.5780 & 3.6032 & 3.6761 & +0.2167 & +0.0729 \\
location, name & 837 & 2.1396 & 2.3612 & 2.3689 & 2.4558 & +0.2216 & +0.0869 \\
company & 1487 & 2.3857 & 2.9034 & 2.9712 & 3.0436 & +0.5177 & +0.0724 \\
company, location, name & 714 & 1.5217 & 1.6181 & 1.6179 & 1.6043 & +0.0964 & -0.0136 \\
\midrule
\texttt{pnav} & 1335 & 1.6293 & 2.0286 & 2.0786 & 2.1458 & +0.3993 & +0.0672 \\
\bottomrule
\end{tabular}
\end{table*}

We test whether segment gains are broadly distributed across query types
rather than driven by a few high-frequency categories. Let
$\mathcal{Q}_g$ denote the queries assigned to category-combination group $g$.
For model $m$, the reported group-conditioned score is
\begin{equation}
\label{eq:per-category-score}
\operatorname{GroupGR}_{m}(g)
=
\frac{1}{|\mathcal{Q}_g|}
\sum_{q\in\mathcal{Q}_g}
\operatorname{FinalGR}_{m}(q).
\end{equation}
We report 21 supported GR category combinations and a separate navigational
query group (\texttt{pnav}); temporal is omitted because the current
category-analysis pipeline does not support it.

Table~\ref{tab:all-category-combinations} reports all 21 GR category
combinations and the separate \texttt{pnav} group. The standalone
segment model outperforms the matched single-vector L0 baseline in all seven
single-category groups. Across the complete analysis, it improves 20 of 21 GR
category combinations and 21 of 22 reported groups when \texttt{pnav} is
included. Its group-macro improvement is $+0.2574$ over the 21 GR combinations
($+0.2639$ including \texttt{pnav}); the only GR-combination regression is
\texttt{location,name,title} ($2.0010\rightarrow1.9452$).

At matched three-slot capacity, the combined segment multi-vector model
outperforms plain tagged-slot MaxSim in every single-category group shown in
Table~\ref{tab:all-category-combinations}. Across the complete analysis, it improves 19 of 21
GR category combinations and 20 of 22 groups including \texttt{pnav}. The
group-macro gain is $+0.0514$ over the GR combinations ($+0.0522$ including
\texttt{pnav}). The two regressions are limited to \texttt{company,name}
($1.7697\rightarrow1.7258$) and \texttt{company,location,name}
($1.6179\rightarrow1.6043$).

These matched-capacity results rule out additional vector count as the sole
explanation for the aggregate improvement. They provide consistent evidence
that the GR-aligned segment design adds value beyond increasing multi-vector
capacity alone, across both single-category and conjunctive multi-category
queries. This group-level consistency complements, rather than replaces, the
paired per-query significance analysis.

\section{Bloom Filter False-Positive Analysis}
\label{sec:fpr_bloomfilter}
\paragraph{Serving context}
First-degree requests carry only $273$ IDs on average ($\approx2.2$\,KB), whereas second-degree sets average about $75{,}000$ IDs and can reach $900$K, precluding direct transmission. We therefore encode large second-degree connection sets in a query-side Bloom filter; the GPU kernel tests each candidate \texttt{User\_ID}, while small sets use exact matching. This keeps the index compact and stateless while decoupling it from connection updates.

The key operational metric is the Bloom filter false-positive probability (FPP), the chance that a non-matching candidate incorrectly tests positive. Even a small FPP matters at corpus scale: on a representative 600M-document corpus, an FPP of 0.1\% yields 600,000 false positives, which can dominate the top-$K$ pool when the query itself returns only 500,000 true positives. Let $\rho$ denote this FPP. The expected candidate volume required to obtain $500$ true matches is $\lceil 500/(1-\rho)\rceil$. Table~\ref{tab:bf-eval} reports this quantity using the displayed FPP values.

\begin{table}[H]
\centering
\footnotesize
\captionsetup{font=small}
\caption{Bloom filter false-positive probability (FPP) and expected candidate volume needed to obtain 500 true matches.}
\label{tab:bf-eval}
\setlength{\tabcolsep}{4.5pt}

\noindent\textit{(a) Bloom filter false-positive probability (\% chance a non-match tests positive)}\par\vspace{0.15em}
\begin{tabular}{@{}lrrr@{}}
\toprule
Second-degree & \multicolumn{3}{c}{Bitmap (KB)} \\
\cmidrule(l){2-4}
connections & 1024 & 2048 & 3072 \\
\midrule
100K & 0.0\% & 0.0\% & 0.0\% \\
200K & 0.0\% & 0.0\% & 0.0\% \\
400K & 0.01\% & 0.00\% & 0.00\% \\
800K & 0.57\% & 0.01\% & 0.00\% \\
1.6M & 14.0\% & 0.66\% & 0.05\% \\
3.2M & 68.0\% & 14.1\% & 2.87\% \\
6.4M & 97.9\% & 67.3\% & 32.1\% \\
\bottomrule
\end{tabular}

\vspace{0.35em}
\noindent\textit{(b) Expected candidate volume to yield 500 true matches}\par\vspace{0.15em}
\begin{tabular}{@{}lrrr@{}}
\toprule
Second-degree & \multicolumn{3}{c}{Bitmap (KB)} \\
\cmidrule(l){2-4}
connections & 1024 & 2048 & 3072 \\
\midrule
100K & 500 & 500 & 500 \\
200K & 500 & 500 & 500 \\
400K & 501 & 500 & 500 \\
800K & 503 & 501 & 500 \\
1.6M & 582 & 504 & 501 \\
3.2M & 1{,}563 & 583 & 515 \\
6.4M & 23{,}810 & 1{,}530 & 737 \\
\bottomrule
\end{tabular}
\end{table}

\subsection{Similarity Scores as a Natural Suppressor}
Semantic embedding similarity acts as an implicit second filter: unrelated false-positive documents rarely score highly against specific queries. Validations using member data show substantially fewer top-$K$ false positives than the conservative, uniform-distribution projections in Table~\ref{tab:bf-eval}. Consequently, a 2\,MB bitmap represents a robust operational sweet spot, maintaining the false-positive probability below $1\%$ for users with up to $\sim1.6$M second-degree connections while adding minimal payload.

\FloatBarrier \section{Full Multi-Vector Serving Sweep}
\label{app:mvr-full-sweep}

Benchmarks use an H200 shard containing ~35M documents with the $256\times3$ treatment.

We swept the re-rank depth $S$ on a infra shards at concurrency 512 (Table~\ref{tab:mvr-perf-full}). At an unsaturated load of 400 QPS, increasing depth tenfold ($2{,}000 \to 20{,}000$) increases P50/P95/P99 latency from 54/72/75\,ms to 123/158/162\,ms. Under a P95 latency target of 150\,ms and a gRPC error-rate limit of 0.5\%, sustainable capacity ranges from 600 QPS at $S=2{,}000$ to 300 QPS at $S=20{,}000$. For intermediate depths of $4{,}000$, $8{,}000$, $10{,}000$, and $16{,}000$, the corresponding capacities are 550, 500, 450, and 350 QPS. Increasing load by only 50 QPS beyond these operating points crosses the saturation cliff, typically raising P95 latency above 300\,ms.

\begin{table}[H]
\centering
\small
\captionsetup{font=small}
\caption{Representative multi-vector operating points on a infra shards. Latency values are in ms; P90 is measured at 400 QPS, and Max QPS satisfies P95 $<150$\,ms with an error rate below $0.5\%$.}
\label{tab:mvr-perf}
\begin{tabular}{@{}lrrr@{}}
\toprule
Depth $S$ & P95@400 & Max QPS & P95@Max \\
\midrule
2{,}000  & 72.8 & 600 & 104.9  \\
4{,}000  & 80.8  & 550 & 112.6 \\
8{,}000  & 104 & 500 & 124.36 \\
10{,}000 & 116.5 & 450 & 134.1 \\
16{,}000 & 151.1 & 350 & 122.1 \\
20{,}000 & 158.74 & 300 & 107.24 \\
\bottomrule
\end{tabular}
\end{table}

\begin{table}[H]
\centering
\footnotesize
\captionsetup{font=small}
\caption{Multi-vector performance benchmarks for second-pass reranking on a infra shard (concurrency 512, $K{=}1000$). All latency values are in ms.}
\label{tab:mvr-perf-full}

\noindent\textit{(a) Latency vs. re-rank depth $S$ at 400 QPS}\par\vspace{0.1em}
\begin{tabular}{@{}lrrr@{}}
\toprule
Re-rank depth $S$ & P50 (ms) & P95 (ms) & P99 (ms) \\
\midrule
2{,}000  & 54.2 & 72.8 & 75.4 \\
4{,}000  & 60.4 & 80.8 & 84.7 \\
8{,}000  & 78.3 & 104.0 & 107.7 \\
10{,}000 & 87.9 & 116.5 & 120.4 \\
16{,}000 & 115.5 & 151.5 & 154.5 \\
20{,}000 & 122.7 & 158.7 & 162.4 \\
\bottomrule
\end{tabular}

\vspace{0.25em}
\noindent\textit{(b) Maximum sustainable capacity (P95 $<150$\,ms; error rate $<0.5\%$)}\par\vspace{0.1em}
\begin{tabular}{@{}lrrrr@{}}
\toprule
Depth $S$ & Max QPS & P90 & P95 & P99 \\
\midrule
2{,}000  & 600 & 102.2 & 104.8 & 107.7 \\
4{,}000  & 550 & 109.8 & 112.6 & 115.5 \\
8{,}000  & 500 & 121.3 & 124.4 & 127.5 \\
10{,}000 & 450 & 130.7 & 134.2 & 137.5 \\
16{,}000 & 350 & 118.8 & 122.2 & 126.1 \\
20{,}000 & 300 & 104.3 & 107.2 & 110.5 \\
\bottomrule
\end{tabular}

\vspace{0.5em}
\scriptsize Observed gRPC error rates at these capacity points range from 0.12\% to 0.33\%.
\end{table}

\section{Fast TopK Implementation and Full Benchmarks}
\label{app:fast-topk}
The kernel executes two passes over the $N$-element FP16 score vector:

\textit{Pass 1 (Threshold Identification):} A histogram is computed over the high byte of each FP16 score using 16-wide vectorized loads (two \texttt{uint4} loads per iteration, 256 bits total, yielding 16 half values). The histogram is reduced device-side to identify the threshold byte $\tau$ containing the $K$-th score, eliminating CPU--GPU synchronization between passes.

\textit{Pass 2 (Fused Collect):} A second pass refines the histogram at byte precision while gathering scores strictly greater than $\tau$ into a pre-allocated output buffer. Boundary values are collected into an overflow buffer and merged device-side to produce exactly $K$ elements. The selected value--index pairs are then sorted on-device, preserving the standard sorted \texttt{topk} interface while sorting only the compact $K$-element result.

Operating natively on FP16 requires only two full passes over the input, compared with four for FP32, limiting full-vector HBM reads to two in the common case. A slow-path fallback re-scans the input when the boundary buffer overflows under pathological tie distributions.

The complete benchmark sweep is reported in
Table~\ref{tab:topk-bench-full}; speedups reach $7.6\times$ over
\texttt{torch.topk} and $6.3\times$ over the chunked baseline.

\begin{table}[t]
\centering
\footnotesize
\captionsetup{font=small}
\caption{Sorted top-$K$ selection latency ($K{=}1000$, FP16, H200 GPU, PyTorch 2.10.0.6, CUDA 12.8). Values are medians of 10 runs after 3 warm-up iterations. The final column reports speedup over \texttt{torch.topk}/Chunked, respectively; $\star$ marks the four target configurations.}
\label{tab:topk-bench-full}
\setlength{\tabcolsep}{2.2pt}
\begin{tabular}{@{}lrrrr@{}}
\toprule
Batch$\times N$ & \texttt{torch.topk} & Chunked & \textbf{Blitz} & Speedup \\
\midrule
$\star\ 1\times50\mathrm{M}$   & 0.59 & 0.57 & \textbf{0.20} & $2.9/2.8\times$ \\
$\star\ 1\times100\mathrm{M}$  & 1.11 & 0.93 & \textbf{0.27} & $4.1/3.4\times$ \\
$1\times200\mathrm{M}$            & 2.27 & 1.65 & \textbf{0.40} & $5.7/4.1\times$ \\
$1\times500\mathrm{M}$            & 5.21 & 3.82 & \textbf{0.80} & $6.5/4.8\times$ \\
$1\times1\mathrm{B}$              & 11.21 & 7.22 & \textbf{1.48} & $7.6/4.9\times$ \\
\midrule
$\star\ 32\times50\mathrm{M}$  & 11.51 & 11.19 & \textbf{3.19} & $3.6/3.5\times$ \\
$\star\ 32\times100\mathrm{M}$ & 26.23 & 25.81 & \textbf{4.68} & $5.6/5.5\times$ \\
$32\times200\mathrm{M}$           & 51.87 & 51.65 & \textbf{8.45} & $6.1/6.1\times$ \\
$32\times500\mathrm{M}$           & 131.10 & 127.26 & \textbf{20.48} & $6.4/6.2\times$ \\
$32\times1\mathrm{B}$             & 261.98 & 255.59 & \textbf{40.53} & $6.5/6.3\times$ \\
\bottomrule
\end{tabular}
\end{table}

\section{Derived Features Benchmark \& Analysis}
\label{sec:appendix_derived_features}

\paragraph{Serving context}
Requests without a network filter previously issued three calls, one with regular attribute filters but no attribute filter, one with first-degree connection filter and one with second-degree connection filter. The dual-role kernel consolidates them into one call and reuses candidate fetches, yielding a $3\times$ efficiency gain at neutral quality. On a 35M-document shard, the attribute-filter-heavy mix sustains 500 QPS at 123.1\,ms P95; with 20\% no-filter requests, it sustains 400 QPS at 124.2\,ms P95.

\begin{table}[!t]
\centering
\small
\captionsetup{font=small}
\caption{Representative derived-features latency under query mixes (33.6M-doc index, model plus two derived features). Latencies are in ms.}
\label{tab:df-latency}
\setlength{\tabcolsep}{3pt}
\begin{tabular}{@{}lrrrr@{}}
\toprule
Query Mix & QPS & P90 & P95 & P99 \\
\midrule
\makecell[l]{100\% attr. filter;\\20\% network filter} & 500 & 120 & 123.1 & 126.1 \\
\makecell[l]{20\% no-filter; 16\% net.-filter;\\64\% other filters} & 400 & 120.6 & 124.2 & 128.9 \\
\bottomrule
\end{tabular}
\end{table}

\subsection{Dual-role kernel implementation}
To execute this efficiently, we implement a \textbf{dual-role kernel} that computes \texttt{compute\_derived\_feature} alongside \texttt{evaluateFilter} within the same GPU thread, reusing the document fetch to eliminate extra memory reads. When an explicit network hard filter is present, the kernel reuses that predicate directly. To mitigate memory write pressure ($F\times N$ booleans per batch), derived features are written exclusively for candidates surviving filtering.


\section{Blinded Human Relevance Evaluation}
\label{app:online-human-eval}
As a study-specific supplementary robustness check, we sample 50 high-frequency queries from top traffic and evaluate Control and Exp on this shared query set, with each model returning 10 ranked results per query (1,000 query--result positions before deduplication). Query volume is capped at 50 by annotation budget. Results from both models were pooled, deduplicated by query--document pair, randomized, and presented without model identity or rank information. Each unique query--document pair received three independent binary relevance judgments from a pool of 20 annotators, and the final label was determined by majority vote. This evaluation was not used for training, model selection, or hyperparameter tuning.

Human P@1 is the majority-vote relevance label of the top-ranked result, and Human P@10 is the fraction of the top 10 results labeled relevant; both are computed per query and then averaged over the 50 queries. Absolute lift is Exp minus Control. We estimate 95\% confidence intervals using 10,000 paired query-level bootstrap samples. Human P@1 uses the exact McNemar test, while Human P@10 uses a paired permutation test.

Table~\ref{tab:human-eval} reports the results. Human P@10 improves by 6.0 points, from 0.61 to 0.67, and is statistically significant ($p=0.004$; 95\% CI $[+2.0,+10.0]$ points). Human P@1 improves by 8.0 points, from 0.68 to 0.76, directionally consistent with the SAGE-judged online P@1 lift (\S6.4), but at $n=50$ this paired difference does not reach significance ($p=0.424$; 95\% CI $[-2.1,+18.0]$ points). We treat P@10 as the primary confirmatory result of this check and P@1 as a directional signal, corroborated by but not independently confirming the online lift.

\begin{table}[H]
\centering
\small
\captionsetup{font=small}
\caption{Blinded evaluation on 50 paired queries. E/C/T denotes Exp wins, Control wins, and ties. Confidence intervals use paired query-level bootstrap; tests are exact McNemar for P@1 and paired permutation for P@10. P@10 is confirmatory here; P@1 is directional at this sample size.}
\label{tab:human-eval}
\setlength{\tabcolsep}{5pt}
\renewcommand{\arraystretch}{1.08}
\begin{tabular}{@{}lcc@{}}
\toprule
Statistic & Human P@1 & Human P@10 \\
\midrule
$N$ queries & 50 & 50 \\
Control & 0.68 & 0.61 \\
Exp & \textbf{0.76} & \textbf{0.67} \\
Absolute lift & +8.0 pp & \textbf{+6.0 pp} \\
Query outcomes (E/C/T) & 9 / 5 / 36 & 31 / 14 / 5 \\
95\% CI for lift & $[-2.1,+18.0]$ pp & $[+2.0,+10.0]$ pp \\
Paired $p$-value & 0.424 & \textbf{0.004} \\
\bottomrule
\end{tabular}
\end{table}

\paragraph{Ethics and privacy.} The A/B experiment followed the company's established experimentation policy. We report only aggregate metrics, and no individual-level member data or identifying information is included in the manuscript. Human annotators were internal participants who followed the company's applicable data-access and handling requirements.

\end{document}